\documentclass{article}

     \usepackage[preprint]{neurips_2018}

\usepackage[utf8]{inputenc} 
\usepackage[T1]{fontenc}    
\usepackage{hyperref}       
\usepackage{url}            
\usepackage{booktabs}       
\usepackage{amsfonts}       
\usepackage{nicefrac}       
\usepackage{microtype}      
\usepackage{natbib}
\usepackage{array}
\usepackage{graphicx}
\usepackage[dvipsnames]{xcolor}
\newcolumntype{L}{>{\centering\arraybackslash}m{3cm}}
\title{Automatic Text Summarization of COVID-19 Medical Research Articles using BERT and GPT-2}

\author{
  Bowen Tan\thanks{BT designed and conducted the pre-processing and quantitative assessment.}\\
  Laboratory of Molecular Genetics\\
  Rockefeller University\\
  New York, NY 10065 \\
  \texttt{btan@rockefeller.edu} \\
  \and
  \textbf{Virapat Kieuvongngam}\thanks{VK proposed the idea, designed the experiment, implemented a prototype model.}\\
  Laboratory of Membrane Biology and Biophysics\\
  Rockefeller University\\
  New York, NY 10065 \\
  \texttt{vkieuvongn@rockefeller.edu} \\
    \and
  \textbf{Yiming Niu}\thanks{YN trained the model, conducted the quantitative assessment.}\\
  Laboratory of Molecular Neurobiology and Biophysics\\
  Rockefeller University\\
  New York, NY 10065 \\
  \texttt{yniu@rockefeller.edu} \\
}

\begin{document}

\maketitle

\begin{abstract}
With the COVID-19 pandemic, there is a growing urgency for medical community to keep up with the accelerating growth in the new coronavirus-related literature. As a result, the COVID-19 Open Research Dataset Challenge has released a corpus of scholarly articles and is calling for machine learning approaches to help bridging the gap between the researchers and the rapidly growing publications. Here, we take advantage of the recent advances in pre-trained NLP models, BERT and OpenAI GPT-2, to solve this challenge by performing text summarization on this dataset. We evaluate the results using ROUGE scores and visual inspection. Our model provides abstractive and comprehensive information based on keywords extracted from the original articles. Our work can  help the the medical community, by providing succinct summaries of articles for which the abstract are not already available.


\end{abstract}

\section{Introduction}

\subsection{COVID-19 Open Research Dataset Chanllenge}
The global health and research community is need of a way to survey the scientific literature to come up with a treatment and measures against the COVID-19. In response to this challenge, the White House and leading research groups have established the COVID-19 Open Research Dataset (CORD-19) to bring in the NLP expertise to help finding the answer within the literature or bringing insights to the public at large \citep{wang2020cord19}. This dataset consists of over 59,000 scholarly articles, including over 47,000 with full text about the COVID-19 or related diseases.

\subsection{Text Summarization}
Automatic text summarization is an active area of research focusing on condensing large piece of text to smaller text retaining the relevant information. There are two general approaches. First is the extractive summarization aiming at extracting and concatenating important span of the source text. This is akin to skimming the text. The second approach focus on generating new summaries that paraphrase the source text. The extractive approach has been shown to maintain reasonable degree of grammaticality and accuracy. On the contrary, abstractive approach is much more challenging, due to the fact that the model must be able to represent semantic information of the source text and then use this semantic representation to generate a paraphrase. However, the model may gain the ability to make a creative use of words or ability to make inference from the source text.

\subsection{Existing Body of Work}
Both approaches of summarization have progressed considerably thanks to the recent advances and the availability of the large pre-trained NLP models, often made use of the attention mechanism \citep{vaswani2017attention}. These models include the Bidirectional Encoder Representations from Transformers (BERT) \citep{devlin2018bert} and more recently OpenAI GPT-2 \citep{noauthororeditor}. They are trained on a very large dataset of text such as the entire corpus of wikipedia and are able to perform well across diverse NLP tasks, including machine translation, question-answering, multiple choice question, text classification, etc. Early text summarization models that uses pre-trained BERT is BERTSUM \citep{liu2019text}. BERTSUM is an extractive modified variant of BERT model trained on a general news (CNN/Daily News) summarization dataset. The model performs binary classification task to predict whether a sentence is to be included in the summary. However, as BERT is not built to perform language generative task, its use for abstractive summarization is limited. 

In the past few years, sequence-to-sequence (seq2seq) models based on the transformer encoder-decoder architecture has been widely used for abstractive summarization \citep{shi2018neural}. From the architecture point of view, the encoder reads the source text and transform it to hidden states and the decoder takes the hidden states and output a summary text. The mapping from hidden representation to output text gives the architecture language generative capability.

Even more recently a unified text-to-text framework has been used to train the large language model on multiple NLP tasks all at once \citep{raffel2019exploring}. The basic idea is to train one single model to map the input text of various tasks to the output text. In this work, we take a similar spirit to fine-tuned a pre-trained OpenAIGPT2 to perform mapping from a selected keywords to a summary text, hence generating a summary abstractly.

\subsection{Low-resource Challenge}
An additional challenge to our task is due to the low availability of the domain-specific corpus. Unlike a more general summarization like the CNN/Daily Mail dataset with 286k document-summary pairs, the COVID-19 related literature, as of April 8 2020, contain approximately 35k full text-abstract pairs. Moreover, the scientific terminology found in the peer-reviewed literature can often be esoteric; thus are not used in the mainstream text where the pre-training was performed. This low-resource may present considerable impediment to the fine-tuning. However, this framework, if found useful, can be further expand.


\section{Approach}

\subsection{Overall Outline}

The project is subdivided into two parts, the unsupervised extractive part is used as a baseline performance and the novel abstractive part. The unsupervised extractive summarization takes the already pre-trained BERT model to perform a sentence embedding; whereby every individual sentence is transformed to 768 high dimensional representation. Subsequently, K-medoid clustering analysis is performed on the high dimensional representation \citep{miller2019leveraging}. Representing semantic centers of text, the cluster centers are selected extracted summary.

Comparing against the extractive summarization, the abstractive summarization is trained to generate a summary from a set of keywords. The keywords are extracted from the source text using existing token classification tools, such as NLTK part of speech tagging packages, or fine-tuned BERT token classifier for part of speech tagging \footnote{Code for BERT token classification is adapted from \url{https://www.depends-on-the-definition.com/named-entity-recognition-with-bert/}. the training dataset is from \url{https://www.kaggle.com/abhinavwalia95/entity-annotated-corpus}}. The keywords are tokens classified as three different groups: verbs, nouns, verbs and nouns. Following the extraction, the keywords are paired with the human-generated abstract (gold summary abstract). This keyword-summary pairs are processed and fed to the GPT-2 model as illustrated in figure 1.

After training, summary results are generated using stochastic sampling method described in section 3.4. The results are compared and qualitative assess by reading inspection. Quantitatively, the generated results are compared against the gold summary using ROUGE score\citep{Lin2004ROUGEAP}.

\subsection{Model Architecture}

Many of the state-of-the-art NLP models are built using transformer architecture \citep{vaswani2017attention}, relying on attention mechanism to convert the input sequences to output sequences. Two kinds of transformer architectures are widely used: the transformer encoder and the transformer decoder. 

The BERT model used here for unsupervised extractive summarization is a pre-trained transformer encoder model \citep{sanh2019distilbert}. The model has 12 attention heads and 6 transformer encoder layers. The output is 768 dimensional last hidden state of the model. We use pytorch-based DistilBERT implemented in the Huggingface transformer  \citep{wolf2019huggingfaces}.

 Due to the GPU resource constraint, the abstractive summarization model is a pre-trained distil version of GPT-2. The DistilGPT2 can take up to 1024 token length. It has 12 attention heads and 6 transformer decoder layers. We use the pytorch version of GPT-2 implemented in the Huggingface transformer package \citep{wolf2019huggingfaces}.

\subsection{Training Strategy of the Abstractive Summarization GPT-2}

The GPT-2 model is trained on 2 tasks: the language modeling (lm) task, and the multiple choice (mc) prediction task. For the lm task, the model predicts a next word token given previous tokens and context. For the mc task, given a set of keywords, the model choose the correct gold summary from summary choices.  Each of the tasks has an associated loss.

The lm task projects the hidden state to the word embedding ouput layer. Cross-entropy loss is applied on the target corresponding to the gold summary to get an lm loss. For the training, we label the start and the end of text with special tokens. To enable the model to recognize the summarization task, a special token, <|summarize|>, is used to separate the keywords and the gold summary. The input are all padded with padding token to 1024 tokens, and any input longer than 1024 tokens are truncated.

For the mc task, the hidden state of the last token, <|endoftext|>, is passed through a linear layer to get a class likelihood score ,i.e. a classification task. The cross-entropy loss is applied to obtain a mc loss. To create the training dataset, we randomly select 3 summaries unrelated to the keywords, so-called distractors, and paired the distractors with the keywords in the similar manners as the gold summary, forming a batch of 4 input items.

The language modeling training labels are the token of summary that are right-shifted by 1 token. This is because GPT-2 is auto-regressive in nature, and the $n^{th}$ token output is generated from all previous n-1 token inputs to the left.

The multiple choice training label is a tensor of a numeric $i$, indicating the $i^{th}$ item that is the correct keyword-gold summary pair.

The total loss is a weighted sum of the two losses at ratio of 2:1 lm loss to mc loss.

\subsection{Intuition of the Training Strategy}

The intuition behind this training strategy is the following. Because the GPT-2 model aims at text generation, it is designed to be auto-regressive. That is to say, the model takes the backward context of n-1 previous tokens to predict the $n^{th}$ token. This is achieved using the masked self-attention mechanism to block the information from tokens to the right of the current position from being calculated. 
The special token <|summarize|> is used to signify the context whereby the subsequent tokens are to be the summary of the information before this special token. The GPT-2 model is to learn this context clue from the fine-tuning.

The multi-loss training is used in the hope that we will be able to induce the model to map the local semantic context in the keywords to the gold summary. At the same time, the model retains the global contextual information of the keywords so that in the end of the text, the model is able to distinguish the gold summary from the distractors. This multi-loss training seem to be prominent in recent language model training that aims at general language understanding \citep{li2019empirical,raffel2019exploring}. 

\begin{figure}[h]
\centering
\includegraphics[width=0.75\textwidth]{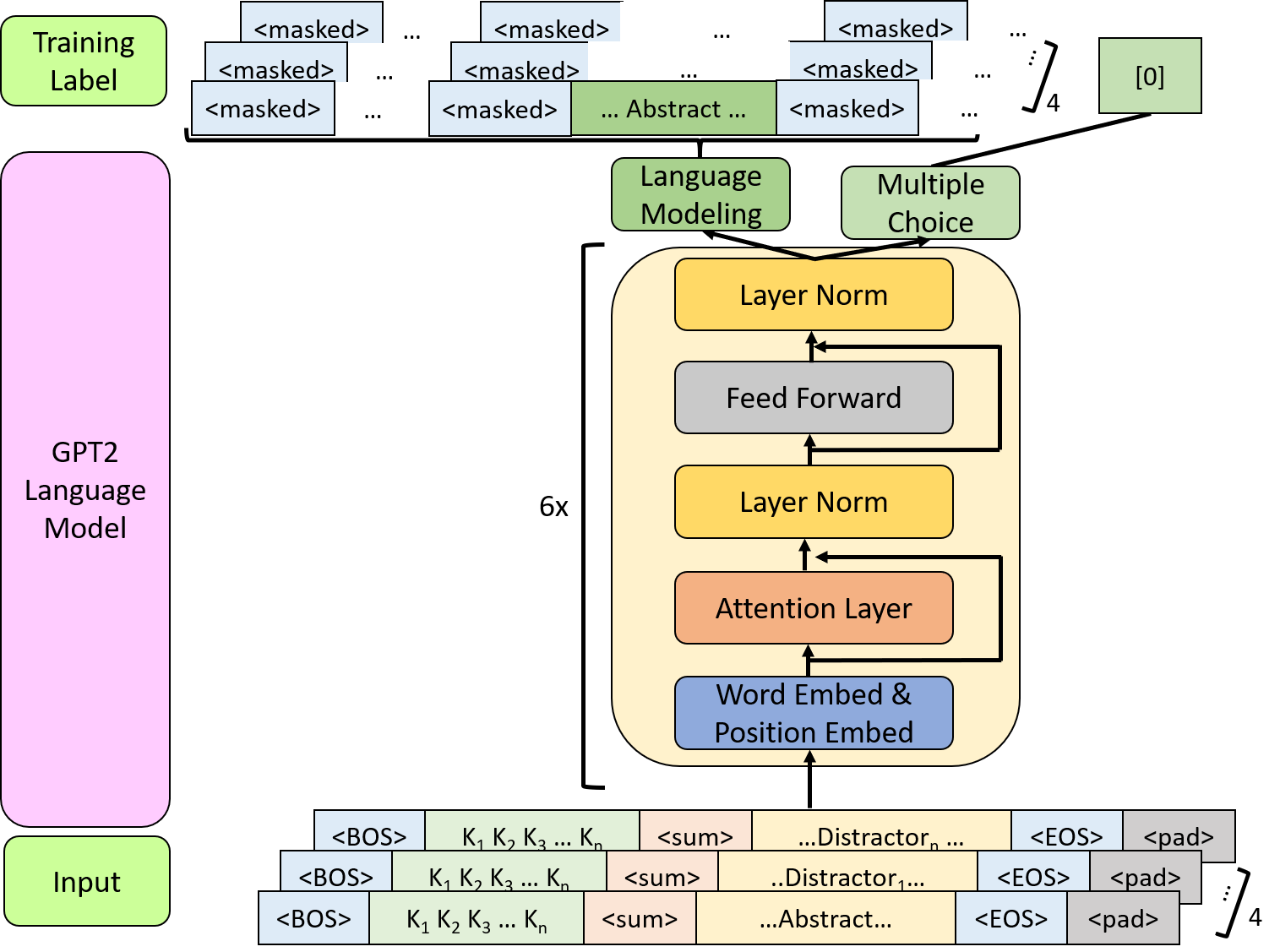}
\caption{\textbf{Overview of GPT2 multi-loss training.} The example input shown has n = 4 items in the input: 1 true abstract and 3 distractors. The true abstract is shown as the $0^{th}$ item. The beginning of sequence, end of sequence, padding, and summarization token are denoted <BOS>, <EOS>, <pad>, <S>, respectively. The language modeling label contains a masked token <masked> everywhere except the gold summary tokens. The multiple choice label is a tensor [0] indicates the $0^{th}$ item as the correct keyword-summary pair.}
\end{figure}

\section{Experiments and Results}

\subsection{Model Training}

The training of DistilGPT2 is carried out on a Google Colab equiped with 1x NVIDIA Tesla P100 GPU. A total of 5 epochs are performed. The training dataset consists of 31246 training samples, each sample has 4 multiple choice options. The validation dataset consists of 3572 samples, each also has 4 multiple choices. The training parameters include the learning rate 3$e^{-5}$, with batch size = 1 and gradient accumulation of 5 steps. The linearly decreasing learning rate scheduler is used for every epoch. The training loss of the first epoch is shown in figure 2. For the validation, the lm loss is 7.0, mc loss is 0, indicating no sign of overfitting.
	
\begin{figure}[h]
\centering
\includegraphics[width=0.8\textwidth]{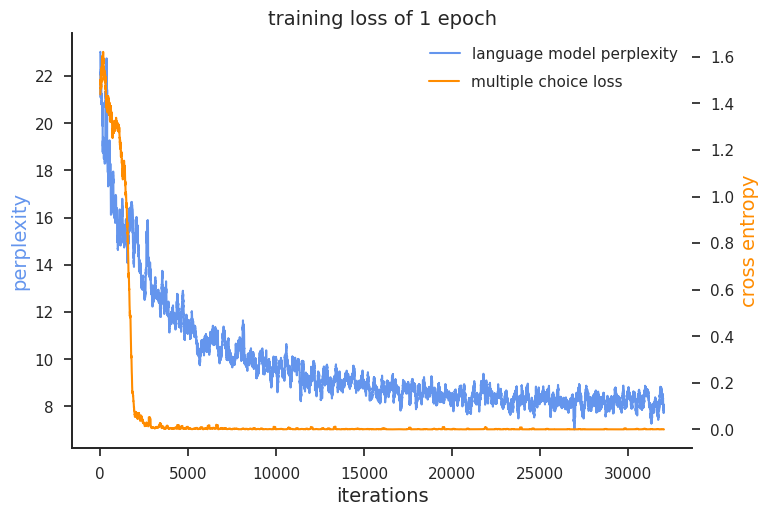}
\caption{\textbf{Training results.} The two losses are shown during 1 epoch of training (32k iterations). The language model loss (blue) is shown in the exponentiated form of the cross-entropy loss, so-called the perplexity score ($e^{lm loss}$). The multiple choice loss (orange) is calculated from the cross entropy loss over all the multiple choices.}
\end{figure}

\subsection{Visualizing the Results}

The attention mechanism allows us to assess the model performance. The attention can be broadly interpreted as a vector of importance weights, i.e. how strongly the tokens in sequences are correlated with other tokens \citep{vig2019transformervis}. To visualize the attention, we input the sequence illustrated in table 1 and plot the attention as matrix of alignment heatmap\citep{bahdanau2014neural}. In the figure 3 below, we visualize the learned attention from the model by comparing the attention of the pre-trained GPT2 before the fine-tuning and after the fine-tuning. Because of the multi-layer, multi-attention head architecture, the total unique structures are 6*12 = 64. We only select 4 attentions that seems to exhibit interesting phenomena. First obvious pattern  exihibited in layer 0 head 1 both before and after fine-tuning is the diagonal pattern. This could be interpreted as representing self-correlation. Observed more strongly only after fine-tuning, the second pattern is the left-shifted diagonal line, shown in layer 3 head 7 and layer 5 head 9. This could be interpreted as the correlation between the keyword input and the summary. This attention, learned during the fine-tuning, became more strongly after more epochs of fine-tuning. This indicates that our training strategy works as expected. Thirdly, the vertical line pattern observed in the attention of layer 2 head 2 both before and after fine-tuning. This could be interpreted as long-range correlation within phrases or sentences in the sequence important for maintaining coherence or grammarticality.

\subsection{Sequence Generation}

The language modeling output of the GPT2 is a tensor of size [sequence length,vocab size]. This is a tensor of likelihood distribution over all words before the softmax. To generate a text sequence from this output, we sample words from this distribution in the word-by-word manner. To obtain the $i^{th}$ word, we consider the conditional probability of previous $i-1$ words.

$P(x) = \Pi_{i=1}P(X_i|x_i,...,x_{i-1})$

Firstly, before the sampling, we can apply a scaling factor, so-called temperature ($t$), to the likelihood ($u$) to reshape the skews likelihood distribution before the softmax \citep{holtzman2019curious}.

$p(x) = \frac{exp(u/t)}{\Sigma_{l^*} exp(u^*_l/t)}$

The high temperature tends to skews the distribution in favor of low probability words; whereas the low temperature skews to the distribution towards a few high probability words. The result is a tug-of-war between favoring generation accuracy at the cost of decreasing word diversity.

Secondly, we employs a stochastic sampling method, called top-p sampling, over the probability distribution of words to pick conditional on previous words in the sequence. The rule for top-p sampling is that the smallest set of candidate words ($x$) to consider is such that the cumulative conditional probability is greater than $p$.

$\Sigma_{x}P(x|x_{1:i-1}) >= p$

Additionally, to prevent the model from sampling too many low probability words, we limit the number of candidate words to consider to $k$ words.

We empirically tested a few of the sampling parameters and found that temperature = 1, k = 50, and p = 0.8 yields a reasonable generations.

\begin{table}[h!]
    \caption{Example of summary result}
    \begin{tabular}{ccc}
    \toprule
        gold summary & keyword input & generated abstractive summary \\
        \midrule
        & 1 epoch training  \\
        \\
        \multicolumn{1}{m{4.5cm}}{Influenza virus is the most frequently reported viral cause of rhabdomyolysis. A 7-year-old child is presented with rhabdomyolysis associated with parainfluenza type 2 virus. Nine cases of rhabdomyolysis associated with parainfluenza virus have been reported. Complications may include electrolyte disturbances, acute renal failure, and compartment syndrome.}&
        \multicolumn{1}{m{4.5cm}}{<|startoftext|>  Rhabdomyolysis Associated with Parainfluenza VirusInfluenza virus is cause rhabdomyolysis child is rhabdomyolysis parainfluenza type 2 virus cases rhabdomyolysis parainfluenza virus have Complications include electrolyte disturbances renal failure compartment syndrome <|summarize|>}&
        \multicolumn{1}{m{4.5cm}}{Influenza virus is the most common cause of respiratory rhabdomyolysis in the child. It is believed that the rhabdomyolysis and parainfluenza type 2 virus cases with rhabdomyolysis in parainfluenza type 2 virus. Recent cases with rhabdomyolysis in parainfluenza virus have been described. Complications include electrolyte disturbances, kidney failure,and gastrointestinal compartment syndrome.
        \newline}\\
        
         &5 epoch training  \\
        \\
        \multicolumn{1}{m{4.5cm}}{}&
        \multicolumn{1}{m{4.5cm}}{<|startoftext|>  Rhabdomyolysis Associated with Parainfluenza', 'VirusInfluenza virus is cause rhabdomyolysis child is rhabdomyolysis', 'parainfluenza type 2 virus cases rhabdomyolysis parainfluenza virus', 'have Complications include electrolyte disturbances renal failure', 'compartment syndrome <|summarize|>}&
        \multicolumn{1}{m{4.5cm}}{Influenza virus is a leading cause of rhabdomyolysis in child. However, several cases of rhabdomyolysis in the parainfluenza virus have been reported. Complications include electrolyte disturbances in renal failure of the normal renal compartment syndrome}
        \\
        \bottomrule
    \end{tabular}
\end{table}

\subsection{Analysis}
 
The result in table 1 in noteworthy in that the model "learns" that influenza virus is \textbf{the most common cause} of rhabdomyolysis (a breakdown of muscle tissue), even though this knowledge is not presented in the keyword. In this respect the model gains an ability to infer some knowledge from the training corpus. However, this could be problematic because it could wrongly infers falsehood as well. For example, the model outputs: complication include ... \textbf{gastrointestinal conpartment syndrome}. In fact, the compartment syndrome is not a gastrointestinal condition. However, we should note that this could also be attributed to our sampling method.

\begin{figure}[h!]
\centering
\includegraphics[width=1\textwidth]{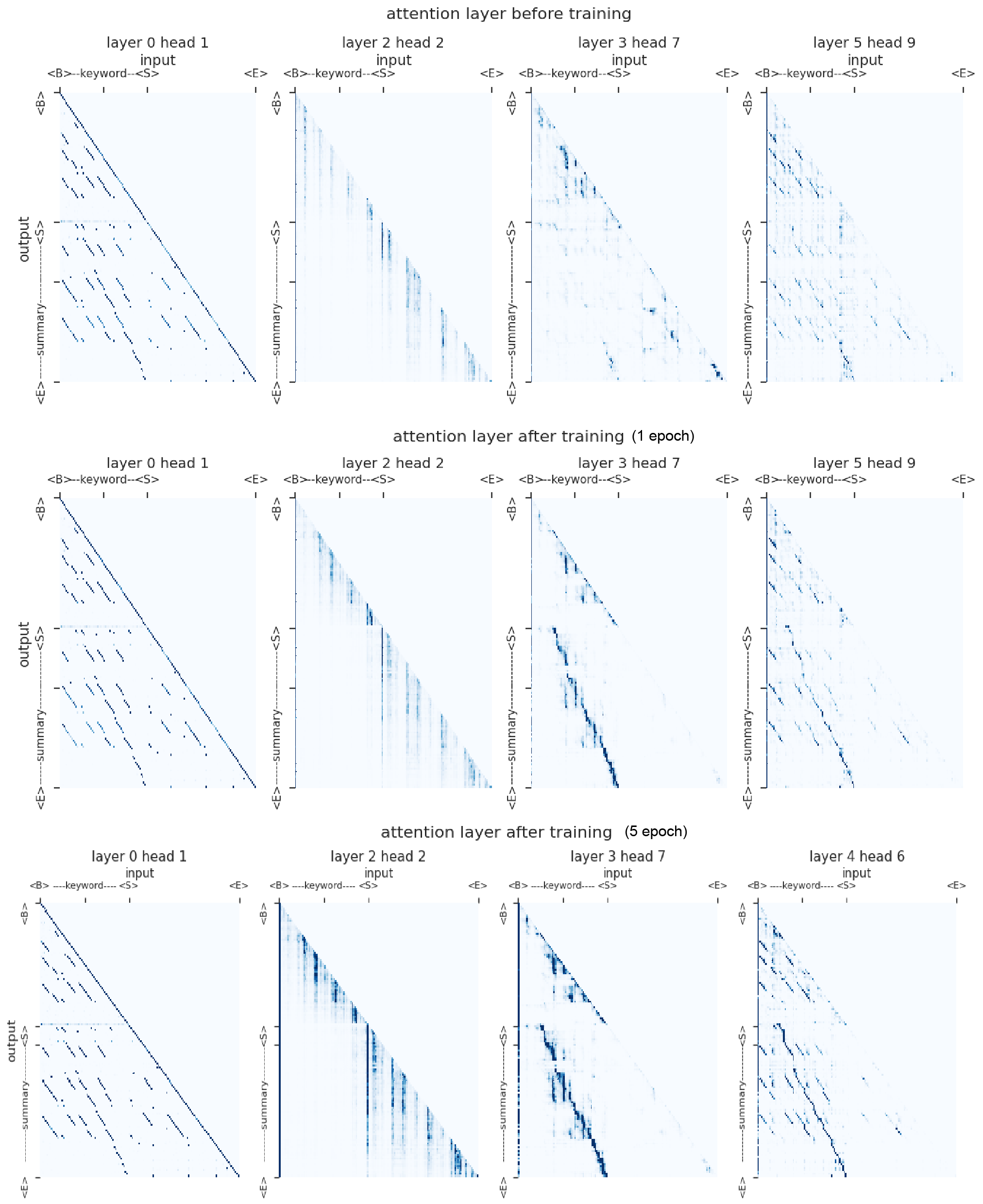}
\caption{\textbf{Visualizing attention mechanism.} The weights of attentions layers mapping the input to the output is shown. The input sequence shown in table 1 is passed through either the pre-trained GPT2 model or our summarization fine-tuned model. The x-axis represents the input sequence. The y axis represents the aligned output. The notation <B>, <S>, <E> denotes the start, summarization, and end token respectively. The part of keyword or summary sequences are labeled. The figure compares of the attention before and after the fine-tuning. Selected attention layers and heads are plotted as matrix heatmaps.}
\end{figure}

\newpage

\section{Quantitative Assessment}

\subsection{ROUGE metric}

ROUGE (Recall-Oriented Understudy for Gisting Evaluation) is a metric used in NLP for evaluating text summarization. The metric compare a model-generated summary against a human-generated reference summary. ROUGE-n measures overlap of n-grams between the two texts. ROUGE-l measures the longest matching sequences of words, without predefined n-gram length, and does not require consecutive matches. ROUGE-w measures the longest matching sequences that take consecutive matching into account \citep{Lin2004ROUGEAP}. The recall version of ROUGE reports the ratio of n-grams in the reference that are also present in the generated summary. The precision version of ROUGE reports the ratio of n-grams in the generated summary that are also present in the reference summary. The F score version of ROUGE is the harmonic mean of the precision ROUGE and recall ROUGE. In our report, we will use the F score version of ROUGE.

\subsection{Extractive Summarization}

We applied kmeans clustering followed by k nearest neighbour to extract sentences representing comprehensive semantic meanings of abstracts. During the extraction, we compared the effects between 40\% versus 60\% compression ratios on ROUGE scores. As shown in figure 4a and 4b, all of the ROUGE scores of 60\% extraction are higher than 40\% extraction, irrespective of training epochs. Additionally, extractive summary produces reuslts with higher rouge scores compared to the abstractive one (Figure 4c). This is consistent with the assumption that less compression during extraction preserves more information compared to the original abstracts. 

\subsection{Abstractive Summarization}

In the abstractive summarization stage, the effect of training time was firstly investigated. As shown in Figure 4, the plotted weights of attentions layers mapping the input to the output i between 1 and 5 epochs suggested a more concentrated distribution. In other words, the GTP-2 model would benefit significantly by longer training process and this is reflected in the better fit of relevance between the input and output. Furthermore, the output summary shown in Table 1 illustrates that longer training could help to generate more 'abstract' summary: Compared to the gold summary, model with 5-epoch training summarize special cases as 'several cases of mixed infections have been reported' rather than explaining the specific case, whereas model with 1-epoch training still tries to describe it. Interestingly, the seemingly more abstract results are not reflected in the calculated rouge score. As shown in Figure 4a, the difference of rouge scores between model with 1 and 5 epochs is insignificant.

To investigate the effect of keywords used for sentence generation, we then compared all of keywords (nouns and verbs) yielded from 40\% versus 60\% extraction. Figure 4a and 4b together show that the overall ROUGE scores from 60\% abstractive group are higher than the 40\% group. This indicates that using more keywords as input tokens would result in more coverage of information compared to original abstracts after GPT2 generator.

We next compared whether different word classes influence generated summaries compared to the original ones. As shown in Figure 4c, the first observation is that 40\% abstraction tend to show lower ROUGE scores than the 60\% group, irrespective of word classes. Second, only using verbs as keywords shows very low ROUGE scores, while only using nouns tend to show almost similar ROUGE scores compared to the group using both verbs and nouns. This may suggest that nouns are generally weighted more than verbs during summary generation, or nouns themselves are representing more accurate information that the original abstracts convey. However, this does not exclude the possibility that this advantage of using nouns is due to using larger percentage of keywords since nouns tend to be used more than verbs in sentences.

In figure 4d, we further evaluate whether using different word sampling methods, the greedy search versus the top-k sampling, would influence the results. Although the ROUGE scores between the two groups are similar, in some cases the greedy search group (top-1) even shows slightly higher scores, the readability and abstractive meanings are significantly worse in the greedy search group compared to the top-k group.

\begin{figure}[h!]
\centering
\includegraphics[width=1\textwidth]{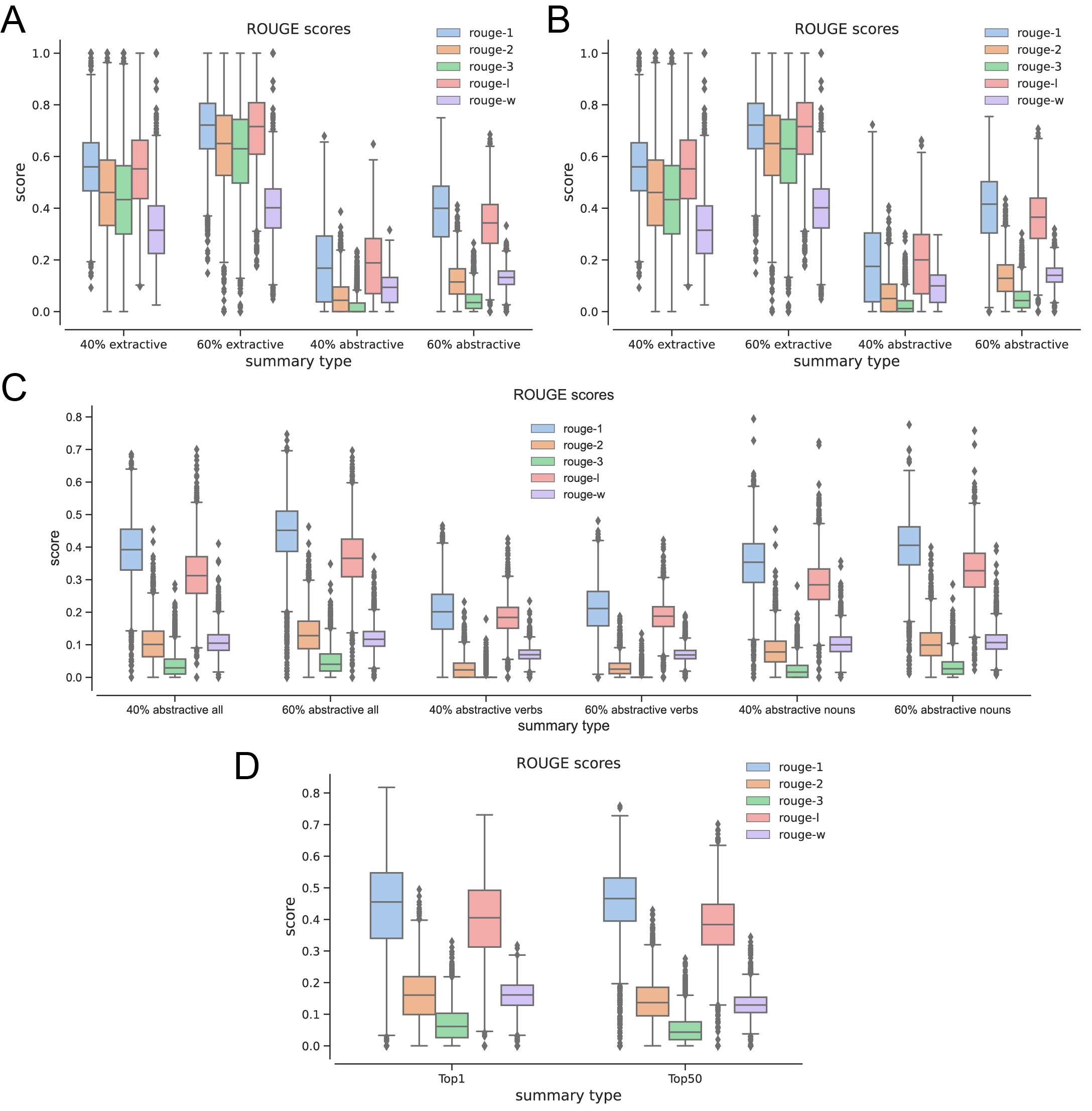}
\caption{\textbf{Summary of experiments on hyper-parameters.} (A) and (B) Comparison of rouge scores between model with 1 (A) and 5 (B) epochs training. (C) Effect of nouns and verbs in keywords. Percentages represent the fraction of specific category of words included. (D) Effect of the greedy search approach. Top 1 and 50 stands for with and without the greedy search, respectively.}
\end{figure}

\section{Conclusion and Future Work}

Abstractive summarization still represents a standing challenge for deep-learning NLP. Even more so when this task is applied to a domain-specific corpus that are different from the pre-training, are highly technical, or contains low amount of training materials. COVID-19 Open Research Dataset Challenge exemplify all the abovementioned difficulty. Nevertheless We have here illustrated that the text-to-text, multi-loss training strategy could be used to fine-tune a pre-trained language model such as GPT-2 to perform abstractive summarization. The result is interpretable and reasonable, even though it is not near human-level performance.

First of all, We think that our model could benefit from further training as the new coronavirus-related research publication are becoming available. This should make the model more accurate in its ability to infer conclusion from the keywords.

In retrospect, we think that the keyword generation phrase is flexible and could also be significantly improved without too much investment. Our keywords are taken from nouns and verbs present in the text. We could investigated how much more fidelity could be gained from adding more information such as adjective part to the keyword. More data augmentation could be performed by randomly dropping or adding words to the keyword sets. This would in effect create much more keyword-summary pairs from the existing ones. Finally, we think that other strategy for extracting keywords could be experimented. For example, one could fine-tune a token classification model to selectively extract scientific keywords.

The evaluation process of abstractive summaries still requires more exploration. From our data, the ROUGE scores represent more direct information or phrases overlapping rather than the actual meanings and readability of summaries. Clearly, all of the extractive models yield higher ROUGE scores than abstractive ones. Intuitively, extractions of raw sentences could result in higher similarity, but this approach would not be favored for better summarization. The fact that the GPT2 generated abstractive summaries showing good readability and succinct information coverage are not reflected by ROUGE scores strongly suggest that other scores or evaluation systems on these aspects are needed in the future. One possible idea would be combining regularization to penalize local similarity but reward global similarity.

Finally, we think that our approach can be further leveraged if more intensive computation resources is available. Overall, our implementation is mostly limited by computation power. With 1 Tesla P-100 GPU, the training can only be done on the DistilGPT2 version with the batch size = 1 because of the memory limit of the GPU. It is likely that the result could greatly benefit from using more bigger GPT-2 version and with more training if permitted by the available computation power. 

In the end, we hope that a text summarization approach. such as ours, can help the medical research community keep up with the rapidly growing literature and that it helps bring new insight to fight the pandemic. 

\section{Code availability}
All source codes and models of GPT2 implemented in this study are publicly available on GitHub \url{https://github.com/VincentK1991/BERT_summarization_1}.


\newpage

\bibliography{biblatex}

\newpage
\section{Appendix}
\subsection{Generated Sample 1}
\textbf{Gold Summary}

\fbox{\begin{minipage}{35em}
Publisher Summary Demyelination is a component of several viral diseases of humans. The best known of these are subacute sclerosing panencephalitis (SSPE) and progressive multifocal leukoencephalopathy (PML). There are a number of naturally occurring virus infections of animals that involve demyelination and many of these serve as instructive models for human demyelinating diseases. In addition to the naturally occurring diseases, many viruses have been shown to be capable of producing demyelination in experimental situations. In discussing virus-associated demyelinating disease, the chapter reviews the architecture and functional organization of the CNS and considers what is known of the interaction of viruses with CNS cells. It also discusses the immunology of the CNS that differs in several important aspects from that of the rest of the body. Experimental models of viral-induced demyelination have also been considered. Viruses capable of producing demyelinating disease have no common taxonomic features; they include both DNA and RNA viruses, enveloped and nonenveloped viruses. The chapter attempts to summarize the important factors influencing viral demyelination, their common features, and possible mechanisms.
\end{minipage}}

\textbf{Abstractive Summary 1}

\fbox{\begin{minipage}{35em}
Abstract Demyelination is a component of many diseases humans are experiencing. Sclerosing panencephalitis (SSPE) and leukoencephalopathy (PML) are common causes of demyelination in many animal and human diseases. However, the virus infections of animals involve demyelination. These situations of demyelinating disease chapter reviews the architecture of the CNS and considers its interactions with the viruses in the CNS. Different aspects of the rest of the body models of demyelination have been explored. Viruses, for example, have been identified as the most important virus infections in humans.
\end{minipage}}

\textbf{Abstractive Summary 2}

\fbox{\begin{minipage}{35em}
Abstract Demyelination is an important component of the diseases human are suffering from. The sclerosing panencephalitis (SSPE) and leukoencephalopathy (PML) are common causes of demyelination in various clinical conditions, and are considered as a potential cause of demyelination in some clinical conditions. However, there are several clinical and epidemiological factors that may influence demyelination in some clinical situations. Demyelination in clinical situations may be a result of a combination of viral and cellular factors, such as viral RNA, viruses, and viruses. In this chapter, we reviews the architecture and organization of the CNS, and considers its potential to be a potential cause of demyelination in some clinical situations. We discuss the immunology of the CNS
\end{minipage}}
\newpage
\subsection{Generated Sample 2}
\textbf{Gold Summary}

\fbox{\begin{minipage}{35em}
RNAi pathway is an antiviral defence mechanism employed by insects that result in degradation of viral RNA thereby curbing infection. Several viruses including flaviviruses encode viral suppressors of RNAi (VSRs) to counteract the antiviral RNAi pathway. Till date, no VSR has been reported in alphaviruses. The present study was undertaken to evaluate chikungunya virus (CHIKV) proteins for RNAi suppressor activity. We systematically analyzed all nine CHIKV proteins for RNAi suppressor activity using Sf21 RNAi sensor cell line based assay. Two non-structural proteins, namely, nsP2 and nsP3 were found to exhibit RNAi suppressor activity. We further validated the findings in natural hosts, namely in Aedes and in mammalian cell lines and further through EMSA and Agrobacterium infiltration in GFP silenced transgenic tobacco plants. Domains responsible for maximum RNAi suppressor activity were also identified within these proteins. RNA binding motifs in these domains were identified and their participation in RNAi suppression evaluated using site directed mutagenesis. Sequence alignment of these motifs across all species of known alphaviruses revealed conservation of these motifs emphasizing on a similar role of action in other species of alphaviruses as well. Further validation of RNAi suppressor activity of these proteins awaits establishment of specific virus infection models.
\end{minipage}}

\textbf{Abstractive Summary 1}

\fbox{\begin{minipage}{35em}
Abstract Chikungunya virus (CHIKV) proteins are highly conserved and are highly conserved in the host cell line. However, their activity against RNAi is not well understood. Here, we evaluated the chikungunya virus (CHIKV) proteins as a non-structural RNAi suppressor activity. We analyzed the CHIKV proteins as a non-structural RNAi suppressor activity. We analyzed the CHIKV proteins as a non- 'structural RNAi suppressor activity. Sf21 RNAi sensor cell line assay proteins nsP2 nsP3 were found to exhibit RNAi activity further. Ourfindings on host Aedes cell lines (EMSA, Agrobacterium infiltration, GFP, tobacco plants, Domains RNAi suppressor activity) were consistent with the observed RNAi suppression mutagenesis. Sequence alignment motifs of species alphaviruses revealed the conserved conservation motifs in the role of action species.
\end{minipage}}

\textbf{Abstractive Summary 2}

\fbox{\begin{minipage}{35em}
Abstract Chikungunya virus (CHIKV) proteins are a novel and promising RNAi-mediated antiviral protein. However, their antiviral activity against chikungunya virus (CHIKV) proteins is not well understood. In this study, we was able to evaluate chikungunya virus (CHIKV) proteins as a novel RNAi-mediated antiviral protein. We detected that CHIKV proteins as a novel RNAi-mediated antiviral protein could suppress chikungunya virus infection by targeting the Sf21 RNAi sensor cell line assay proteins nsP2 and nsP3. We were able to demonstrate that CHIKV proteins as a novel RNAi-mediated antiviral protein could suppress chikungunya virus infection by targeting the Sf21 RNAi sensor cell line assay proteins nsP2 and nsP3
\end{minipage}}

\end{document}